\documentclass[a4paper]{article}

\usepackage{INTERSPEECH2021}

\title{An Investigation of the Relation Between Grapheme Embeddings and Pronunciation for Tacotron-based Systems}
\name{Antoine Perquin$^1$
, Erica Cooper$^2$, Junichi Yamagishi$^2$}

\address{$^1$Univ Rennes, CNRS, IRISA, France 
	$^2$National Institute of Informatics, Japan}
\email{antoine.perquin@irisa.fr, ecooper@nii.ac.jp, jyamagis@nii.ac.jp}

\begin{document}

\maketitle

\begin{abstract}
End-to-end models, particularly Tacotron-based ones, are currently a popular solution for text-to-speech synthesis. They allow the production of high-quality synthesized speech with little to no text preprocessing. Indeed, they can be trained using either graphemes or phonemes as input directly. However, in the case of grapheme inputs, little is known concerning the relation between the underlying representations learned by the model and word pronunciations. This work investigates this relation in the case of a Tacotron model trained on French graphemes. Our analysis shows that grapheme embeddings are related to phoneme information despite no such information being present during training. Thanks to this property, we show that grapheme embeddings learned by Tacotron models can be useful for tasks such as grapheme-to-phoneme conversion and control of the pronunciation in synthetic speech.
\end{abstract}


\section{Introduction}
\label{sec:intro}

The resurgence of neural networks has significantly changed the traditional pipeline used for text-to-speech (TTS) synthesis. At first, they were integrated in the signal generation process by replacing the the statistical model of statistical parametric systems \cite{ze2013statistical} or to define the target cost for unit selection \cite{wan2017google}. Then, each module of the pipeline was replaced by neural solutions \cite{arik2017deep}. The aim of end-to-end TTS systems is to ultimately replace the whole pipeline by a single neural network predicting the audio signal corresponding to the reading of a given text. Meanwhile, current end-to-end systems use a neural network to predict an acoustic representation of speech from text, then convert the acoustic sequence into actual speech by using a neural vocoder. Such end-to-end systems include Tacotron \cite{wang2017tacotron}, which has been shown to outperform traditional approaches when paired with a high-quality vocoder \cite{shen2018natural}.

The original Tacotron model used graphemes as input, which allows the text processing pipeline to remain as simple as possible. However, most following works used phonemes instead. Despite increasing the complexity of the pipeline, the advantage is three-fold. (1) It allows the avoidance of potential pronunciation errors induced by the model mis-learning how to pronounce certain words \cite{skerry2018towards} and makes neural network training easier. (2) It also allows training of the model on languages like Chinese and Japanese \cite{yasuda2019investigation}, where learning pronunciation from graphemes is complicated since their orthographies contain ideograms. (3) The phoneme input allows users to phonetically control synthesized speech. This is useful when the users want to create and use personalized pronunciation dictionaries for certain words.

The performance of grapheme-based and phoneme-based sequence-to-sequence models has been compared with conflicting results for the English language. In \cite{fong2019comparison}, the grapheme-based model performs significantly worse than the phoneme-based one for naturalness. On the contrary, \cite{yasuda2020investigation} found that both types of inputs led to similar results if the model contained a CBHG module (which is the case for Tacotron-based models). 
In this work, we aim to analyze how the embedded representations of a Tacotron system trained on graphemes relate to pronunciation. Before this investigation, in order to assess the quality of the trained model, we show that in the case of a well-curated French dataset, both grapheme and phoneme inputs lead to equivalent performance in terms of pronunciation and overall quality.

Using this well-trained Tacotron model with French grapheme input, this paper conducts an analysis of the embedded representations of Tacotron's encoder. When a Tacotron model is trained on grapheme inputs, the learning process for word pronunciation is done without the supervision of any phonemic knowledge. There is currently little to no knowledge on the kind of representations learned by Tacotron models from grapheme inputs. Can the network automatically learn a linguistically-motivated representation of text, such as phonemes? Alternatively, are those representations completely different from phonemes but still suitable for acoustic feature generation? If the embedded representations are well-correlated with phonemes, it may be possible to phonetically control synthesized speech even if we use graphemes as inputs. In order to answer this question, we visualize Tacotron's embedded representations, analyze phoneme prediction performance from the embeddings, and conduct a proof-of-concept experiment to control pronunciations of the grapheme-based Tacotron.

\section{Tacotron-based Model Architecture}
\label{sec:model_and_data}

\begin{figure}
    \begin{center}
	\includegraphics[clip, trim=0cm 2.8cm 0cm 2.8cm, width=0.65\columnwidth]{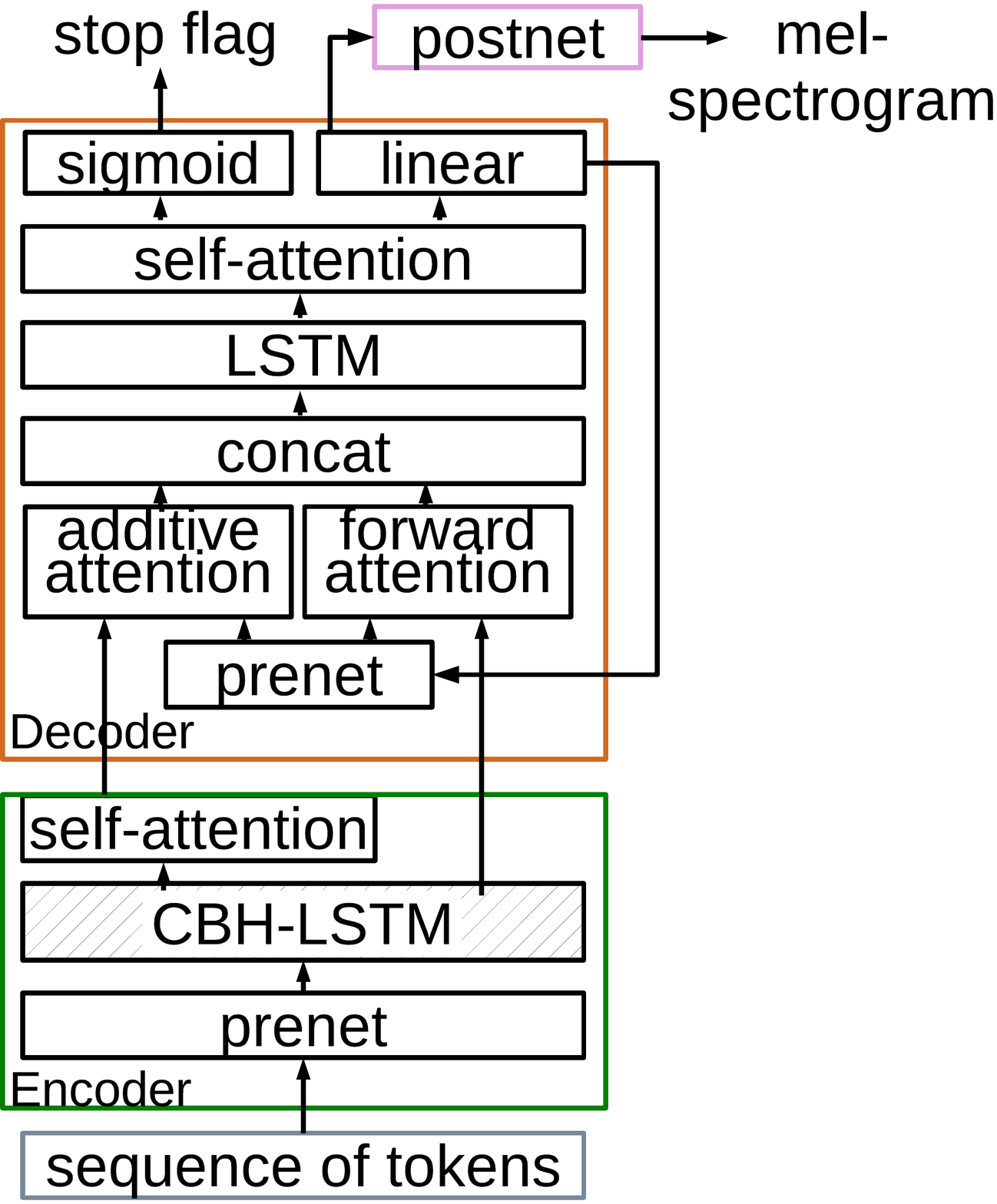}
	\caption{Architecture of the neural network. \textit{The ouput of the highlighted CBH-LSTM module is used to extract the embeddings analysed in Section \ref{sec:embeddings_analysis}}.}
	\label{fig:tacotron_model}
	\end{center}
\end{figure}

The TTS system under study is composed of a Tacotron-based model to predict mel-spectrograms from textual data, and a WaveRNN \cite{kalchbrenner2018efficient} model to convert the mel-spectrograms into speech waveforms. The diagram of the system is shown in Figure \ref{fig:tacotron_model}.  A minor difference between the original Tacotron and our model is the addition of a self-attention block \cite{vaswani2017attention} in the encoder and decoder. This block is added after the CBHG module, and the outputs of both are used as the encoder prediction. The decoder treats each of these with a different type of attention. The output of the encoder CBHG module is attended over with forward attention, while the output of the encoder's self-attention module is attended over with additive attention. Finally, self-attention is also added in the decoder after the decoder recurrent layer. These modifications follow those in  \cite{yasuda2019investigation} and allow the capturing of long term information while speeding up the alignment process. 

We use the SIWIS \cite{honnet2017siwis} dataset to train our models. It contains high-quality speech recordings of a French female professional voice actress reading mainly sentences from parliamentary sessions and French books. The parliamentary sentences were selected according to phoneme coverage and the book sentences were selected to cover the 10,000 most frequent words in the French language. 
The text was cleaned to expand common abbreviations and to spell out numerals. Phoneme transcriptions of the texts are extracted using the phonetizer of the Espeak\footnote{http://espeak.sourceforge.net/} synthesizer, which is dictionary based, with hand-crafted rules to deal with unknown words. 
The audio is downsampled to 24kHz, and 80-dimensional mel-spectrograms are extracted after beginning and trailing silences were automatically trimmed. In total, around 9 hours of speech and 8,000 sentences are preprocessed. The data are split into train/dev/test sets by leaving out 200 and 400 sentences for the test and dev set respectively. Two models are trained with either one-hot encoded text or phoneme sequences as input. 
A WaveRNN model is also trained on the training set.

\section{Experiments}
\label{sec:experiments}

Before analyzing the relation between embedded representations of our Tacotron system and pronunciation, we first need to assess its quality. To do so, we compare our model trained on graphemes to a baseline trained on phonemes.

\subsection{Listening Test}
\label{sub:listening_test}

Out of the 200 test sentences, 50 are randomly chosen to be synthesized by both the grapheme and phoneme models; analysis by synthesis using the WaveRNN neural vocoder is also done. We ran a listening test to compare both models on quality and pronunciation.\footnote{Audio samples and pretrained models are available at \\ https://nii-yamagishilab.github.io/samples-french-tacotron
}

In a MUSHRA-like manner, 4 samples (natural, analysis-synthesis, grapheme model, phoneme model) are randomly presented on a single page. Listeners are asked to evaluate the overall quality of the samples with an opinion score between 0 and 100, where the scale is annotated from ``Bad (0-20)" to ``Excellent (80-100)". As a proxy for a full intelligibility test, the listeners are also asked to note the level of pronunciation errors compared to a reference text with a score between 1 and 5, each point being annotated with: ``1: Does not match the text; 2: Annoying; 3: Slightly annoying; 4: Perceptible but not annoying; 5: Imperceptible.'' In total, 12 French native speakers took part in the listening test. 


\begin{figure}
    \begin{center}
	\includegraphics[width=1\columnwidth]{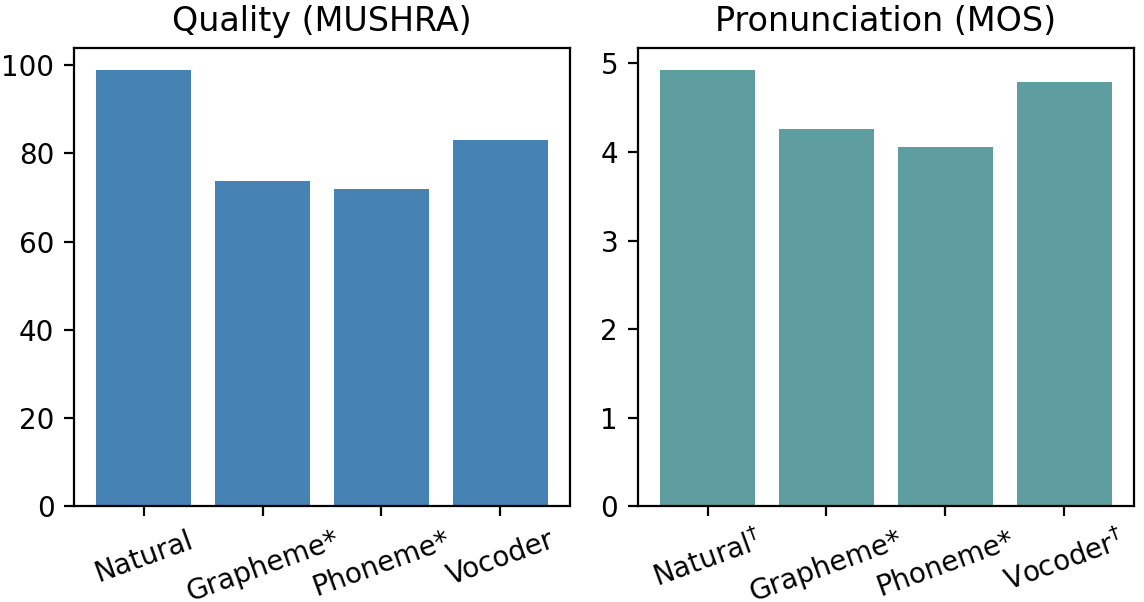}
	\caption{Listening test results.  Within the same graph, differences that are \textbf{not} statistically significant ($p>0.05$) are indicated by pairs of the symbols * or $^\dagger$.}
	\label{fig:listening_test}
	\end{center}
\end{figure}

The results of the listening test are shown in Figure \ref{fig:listening_test}.  For the overall quality, samples synthesized with analysis by synthesis are rated with an averaged MUSHRA score of 83, which indicates that the WaveRNN was trained successfully but suffers from a big dip in quality compared to natural samples. The grapheme and phoneme based models are respectively rated with a score of slightly over 70, which corresponds to a ``good" quality according to our MUSHRA scale. The difference between the two values is not statistically significant according to a Wilcoxon signed-rank test, which suggests that using graphemes directly instead of phonemes does not lower the overall quality of synthesized speech.

For the pronunciation scores, both the natural and vocoded samples were rated with MOS close to 5. Since the SIWIS dataset corresponds to the recording of a professional voice actor, the natural samples do not contain pronunciation errors, and the vocoding process does not create any. Both the grapheme and phoneme models were rated with MOS slightly higher than 4, which indicates that most of the pronunciation errors that occur are not annoying to listeners. We found that most pronunciation errors were not mispronounced words but missed or superfluous liaison\footnote{For example, the ``\textit{\underline{s}}" is pronounced /z/ in ``\textit{le\underline{s} enfants}" due to liaison}. Again, the difference between the two models is not statistically significant. This suggests that using graphemes instead of phonemes does not impact the amount of pronunciation errors, either.

Overall, training the model on graphemes or phonemes leads to similar quality and pronunciation errors. As such, the grapheme model is suitable to use for performing an analysis of the underlying embeddings and how they relate to pronunciation. This study will be the subject of Section \ref{sec:embeddings_analysis}.

\subsection{Objective Evaluation of Pronunciation Errors}
\label{sub:objective_evaluation_ASR}

To further investigate the difference in pronunciation errors between the grapheme and phoneme models, we evaluated them objectively using Automatic Speech Recognition (ASR) on their synthesized outputs. We used the IBM Watson ASR service\footnote{https://www.ibm.com/cloud/watson-speech-to-text} to obtain transcripts of our audio samples. To limit the impact of errors due to the language model, we extracted phoneme sequences for both the transcripts and reference sentences using Espeak. The phoneme sequences were then compared by  Levenshtein distance. The Phoneme Error Rate is computed as the number of insertions, deletions and substitutions needed to transform the predicted transcript into the reference sequence, divided by the number of phonemes in the reference sequence. We performed the experiment on two datasets:\ the 50 sentences used in the listening test, and a compilation of 228 French tongue twisters (with no audio reference)\footnote{https://www.lawlessfrench.com/pronunciation/tongue-twisters/}. By evaluating the amount of pronunciation errors on tongue twister sentences, we study a use-case where the sentences to synthesize differ largely from the training data. This allows us to evaluate how well the grapheme-based model has learned to read a text in an edge case.

\begin{table}[tb]
    \caption{Phoneme Error Rate [\%] of all models after automatic speech recognition and phonetization of the transcripts.}
    \small 
	\label{tab:PER_SIWIS}
	\begin{tabular}{|c|c|c|c|c|}
		\hline
		& Natural & Grapheme & Phoneme & Vocoder \\ \hline
		SIWIS sentences & 6.10* & 7.82*& 10.98* & 8.90* \\ 
		Tongue twister & / & 23.37$^\dagger$ & 24.00$^\dagger$ & / \\ \hline
	\end{tabular}
\end{table}

\begin{figure*}
	\centering
	\includegraphics[width=0.6\linewidth]{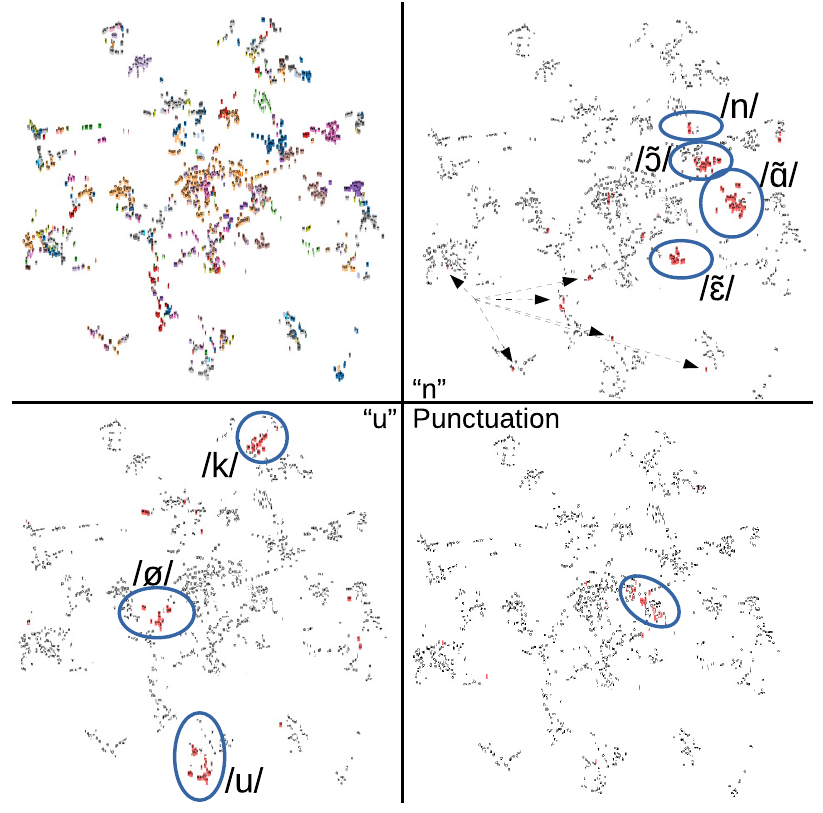}
	\caption{Visualization of Tacotron's embeddings at the encoder. Top-left:\ every embedding with a color per grapheme. Top-right:\ highlights ``n" graphemes. Bottom-left:\ highlights ``u" grapheme. Bottom-right:\ punctuation.}
	\label{fig:characterembeddings}
\end{figure*}

The results are presented in Table \ref{tab:PER_SIWIS}. Interestingly, the difference between the samples synthesized from the grapheme and phoneme models is not statistically significant for both the types of texts. This is consistent with the results of the listening test: using graphemes instead of phonemes does not seem to increase the amount of pronunciation errors in synthesized speech.
For tongue-twister sentences, the PER increases to slightly more than 20\%, which is not surprising since those sentences are harder to synthesize and to comprehend.

\section{Analysis of Tacotron's Embeddings}
\label{sec:embeddings_analysis}

As seen in Section \ref{sec:experiments}, a Tacotron model can be successfully trained on textual data in the form of graphemes instead of phonemes, without losing any quality or increasing pronunciation errors. This suggests that the model learns to read from graphemes. The aim of this section is to explore what sort of representations are learned for that task.

\subsection{Visual Observation}
\label{sub:visual_observation}

We focus on the embeddings at the output of the LSTM layer at the end of the CBHG module in the encoder (highlighted in Figure \ref{fig:tacotron_model}). All sentences of the test set are fed to the grapheme model to extract the corresponding embeddings. They are then visualized using t-SNE \cite{maaten2008visualizing} to obtain Figure \ref{fig:characterembeddings}. For visualization purposes, the number of points was restricted to 3,000 embeddings chosen randomly.

Every grapheme is represented by a different color. Since there was no alignment between graphemes and phonemes, we relied on the words a given grapheme appeared in to manually deduce the corresponding phoneme for this analysis. From the t-SNE, groups distinguish themselves by their position, some containing only one type of grapheme, others mixing them. This suggests that the Tacotron model does not rely solely on individual grapheme labels to deduce pronunciation.

The second picture highlights the position of ``n" graphemes in the embedding space. By their distribution, multiple clusters of the ``n" embeddings can be found. After analysis, each cluster can be associated to a different phoneme, such as /n/ or /\~a/. Interestingly, the clusters do not contain embeddings corresponding to a unique sequence of graphemes. For exemple, the /\~a/ group is made of embeddings corresponding to grapheme sequences such as ``ans", ``anc'', or ``ent". More surprisingly, some ``m" graphemes appear in the /\~a/ cluster. They correspond to sequences such as ``emb" which also corresponds to the /\~a/ phoneme. Finally, some grapheme embeddings for ``n" do not belong to any cluster, as indicated by the dashed arrows.

The same observations can be made for vowel graphemes such as ``u". Groups of embeddings can be found which actually correspond to a single phoneme, and some other graphemes forming the same phoneme can be found in those clusters. The fourth picture highlights punctuation such ``?", ``.", ``?" or ``,". All of the corresponding embeddings are found grouped together. Some ``s" graphemes that appear in sequences of graphemes where the ``s" is muted can be found in the same cluster as the punctuation graphemes. This suggests that all punctuation marks are seen as similar by the model, and correspond to a silent pronunciation.

Overall, this visual observation gives interesting insights into the structure of the representation learned by the Tacotron model. First, this machine-learned representation can be interpreted. Second, Tacotron's embeddings at the encoder are not distributed solely according to the corresponding grapheme, but take its context into account. Third, sequences of different graphemes corresponding to the same phoneme are grouped together. This suggests the model learned to rely on the same distinctive unit suggested by linguistics: the phoneme.

\subsection{Tacotron as Grapheme-to-Phoneme Conversion}
\label{sub:g2p}

We found that Tacotron models are able to extract representations akin to phonemes from grapheme texts. This is not surprising, considering the CBHG module contains a filter-bank. It allows the consideration of grapheme sequences of different lengths, similar to N-grams. It is then able to learn statistically which phoneme is most probable for each sequence of graphemes, similar to statistics-based grapheme-to-phoneme (G2P) converter. This suggests Tacotron's embedded representations might be used for G2P conversion.

To assert this hypothesis, we build a simple shallow model to perform phoneme prediction from Tacotron embedded representations using the ESPnet toolkit \cite{inaguma-etal-2020-espnet}. The model is composed of a single LSTM layer with 64 units, using CTC \cite{graves2006connectionist} as a loss. The CTC loss allows the model to make labeled predictions from unaligned data. In our case, this allows us to train the model without having aligned phoneme and grapheme sequences. The model is trained for 20 epochs on either grapheme-based Tacotron's embeddings or one-hot encoded raw graphemes. The training data come either from the training set or validation set defined in Section \ref{sec:model_and_data}. The precision of the models is measured on the test set using PER as in Section \ref{sub:objective_evaluation_ASR}. The results are reported in Table \ref{tab:PER_G2P}.

When trained on a large amount of data (train set), the G2P model trained on Tacotron's embeddings gets a mean PER value of 13\%. The precision of this model is better than the one trained directly on graphemes, and the difference is statistically significant. This suggests that Tacotron models extract an embedded representation that is closer to phonemes than graphemes. When trained on a much smaller amount of data (the dev set contains only 400 sentences), the model trained on Tacotron's embeddings suffers a slight loss in precision. This suggests that the model could benefit from more training data for optimal results. This could also mean that embeddings extracted from data unseen during the training of the Tacotron model are slightly less related to phonemes. In the case of the model trained on raw graphemes, the PER increases substantially. The high gap in precision between the two models trained on the dev set leads to an interesting use case: Tacotron's embedded representations might be used to train G2P models for low-resource languages. We will investigate this in our future work. 

\begin{table}[tb]
	\caption{Phoneme Error Rate [\%] of G2P models learned from Tacotron's embedded representations or raw graphemes.}
	\label{tab:PER_G2P}
	\small
	\centering
	\begin{tabular}{|c|c|c|}
		\hline
		& Raw graphemes & Tacotron embeddings \\ \hline
		Train set & 19.3 & 12.8 \\ 
		Dev set   & 30.2 & 16.4 \\ \hline
	\end{tabular}
\end{table}

\subsection{Phoneme Control in Grapheme based Tacotron}

Contextual grapheme embeddings encode phoneme information, so they should be usable to control the pronunciation of synthesized speech. As a proof of concept, we extracted the contextual grapheme embedding sequences of two sentences and switched the embedding of a ``b" grapheme in the first sentence by that of an ``m" grapheme in the second sentence before resuming synthesis. When the context of the ``m" and ``b" graphemes are similar, the synthesized phoneme is recognized as /m/ instead of /b/. When the contexts are largely dissimilar, listeners are unsure about the synthesized phoneme. Further work needs to be done to investigate the use of grapheme embeddings for phonemic control. Synthesized samples are available on the listening page.

\section{Conclusions} 
We trained a Tacotron model directly on graphemes rather than phonemes and showed that in the case of a well-curated French dataset, the system performs similarly to a system trained on phonemes. Such a system embeds graphemes in a way that encodes phoneme information. Thus, Tacotron's embedded grapheme representations might be used for a wide variety of applications such as grapheme-to-phoneme or control of the synthesized pronunciation.

In this work, we focused on the French language. We believe our analysis should hold true for other alphabet-based languages such as English. Indeed, neither language have a phonemic orthography, ie the mapping between grapheme and phoneme is not one-to-one. However, the consistency of orthography varies between languages. For exemple, French's orthography is more consistent than English, but less consistent than German \cite{aro2003learning}. Thus, future works should evaluate the impact of a language's orthography consistency on our findings. 

Our future work will also include more detailed investigations of phoneme control in grapheme-based Tacotron. 

\section{Acknowledgements}

This work was partially supported by a JST CREST Grant (JPMJCR18A6, VoicePersonae project), Japan,  MEXT KAKENHI Grants (16H06302, 18H04120, 18H04112, 18KT0051, 19K24372, 19K24373, 19K24371), Japan.  The numerical calculations were carried out on the TSUBAME 3.0 supercomputer at the Tokyo Institute of Technology.

\bibliographystyle{IEEEtran}

\bibliography{mybib}

\end{document}